\documentclass[dvipsnames,format=sigconf]{acmart}
\AtBeginDocument{%
  }
    
\usepackage{algorithm}
\usepackage{algorithmic}
\usepackage[inline]{enumitem}

\acmDOI{10.1145/nnnnnnn.nnnnnnn} % To be updated after completing copyright process
\acmISBN{978-x-xxxx-xxxx-x/YY/MM} % To be updated after completing copyright process
\acmConference[GECCO '26]{The Genetic and Evolutionary Computation Conference 2026}{July 13--17, 2026}{San José, Costa Rica}
\acmYear{2026}
\copyrightyear{2026}

\begin{document}

\title{Distributional Value Estimation Without Target Networks for Robust Quality-Diversity}

\author{Behrad Koohy}
\email{behrad.koohy@luffy.ai}
% \orcid{XXXX-XXXX-XXXX-XXXX}
\affiliation{%
  \institution{Luffy.AI}
  \city{Culham}
  \state{Oxfordshire}
  \country{United Kingdom}
}

\author{Jamie Bayne}
\email{jamie.bayne@luffy.ai}
% \orcid{XXXX-XXXX-XXXX-XXXX}
\affiliation{%
  \institution{Luffy.AI}
  \city{Culham}
  \state{Oxfordshire}
  \country{United Kingdom}
}

\renewcommand{\shortauthors}{Koohy et al.}
%%
%% Article type: Research, Review, Discussion, Invited or position
\acmArticleType{Research}
%%
%% Links to code and data
\acmCodeLink{{\color{red} https://github.com/borisveytsman/acmart}}
% \acmDataLink{htps://zenodo.org/link}
%%
%% Authors' contribution
% \acmContributions{BT and GKMT designed the study; LT, VB, and AP
%   conducted the experiments, BR, HC, CP and JS analyzed the results,
%   JPK developed analytical predictions, all authors participated in
%   writing the manuscript.}
%%
%% Sometimes the addresses are too long to fit on the page.  In this
%% case uncomment the lines below and fill them accodingly.
%%
%% \authorsaddresses{Corresponding author: Ben Trovato,
%% \href{mailto:trovato@corporation.com}{trovato@corporation.com};
%% Institute for Clarity in Documentation, P.O. Box 1212, Dublin,
%% Ohio, USA, 43017-6221}

\begin{abstract}
Quality-Diversity (QD) algorithms excel at discovering diverse repertoires of skills, but are hindered by poor sample efficiency and often require tens of millions of environment steps to solve complex locomotion tasks. Recent advances in Reinforcement Learning (RL) have shown that high Update-to-Data (UTD) ratios accelerate Actor-Critic learning. 
While effective, standard high-UTD algorithms typically utilise target networks to stabilise training. This requirement introduces a significant computational bottleneck, rendering them impractical for resource-intensive Quality-Diversity (QD) tasks where sample efficiency and rapid population adaptation are critical.
In this paper, we introduce QDHUAC, a sample-efficient, target-free and distributional QD-RL algorithm that provides dense and low-variance gradient signals, which enables high-UTD training for Dominated Novelty Search whilst requiring an order of magnitude fewer environment steps.
We demonstrate that our method enables stable training at high UTD ratios, achieving competitive coverage and fitness on high-dimensional Brax environments with an order of magnitude fewer samples than baselines. Our results suggest that combining target-free distributional critics with dominance-based selection is a key enabler for the next generation of sample-efficient evolutionary RL algorithms.

\end{abstract}

\keywords{Genetic Algorithms, Quality-Diversity, Actor-Critic, Reinforcement Learning}

\begin{teaserfigure}
  \centering
  \includegraphics[width=\textwidth, trim={0cm 0.25cm 0cm 0.25cm}, clip]{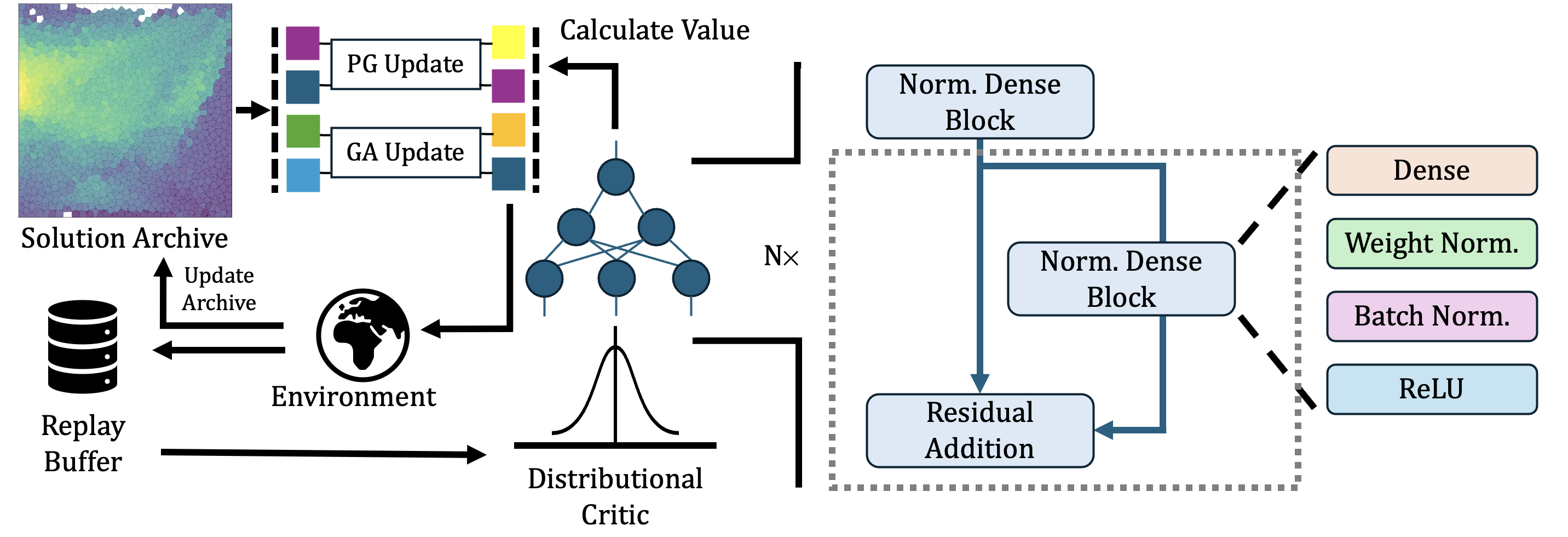}
  \caption{\textit{The Optimisation Loop of QDHUAC.} The algorithm maintains a shared Replay Buffer $\mathcal{D}$ populated by the evolutionary search. A global Target-Free Distributional Critic learns continuously from this diverse data to guide a gradient-based policy, which accelerates the discovery of high-performing solutions. To stabilise learning at high Update-to-Data (UTD) ratios without a target network, the critic employs compact residual blocks utilising Hybrid Normalisation. We apply Weight Normalisation to control the gradient magnitude, followed by Batch Normalisation to mitigate the internal covariate shift caused by the non-stationary archive distribution. This architecture keeps the target-free critic and the population synchronised.}
  \label{fig:teaser}
\end{teaserfigure}

\maketitle

% \begin{figure*}
%     \centering
%     \includegraphics[width=0.95\linewidth]{assets/loop.png}
%     \caption{Caption}
%     \label{fig:placeholder}
% \end{figure*}

\section{Introduction}
Quality-Diversity (QD) optimisation focuses on the discovery of diverse and high-performing skill repertoires that solve complex tasks in parallel. 
% By combining evolutionary search with local competition, algorithms such as MAP-Elites \cite{mouret2015illuminating} and Dominated Novelty Search (DNS) \cite{bahlous2025dominated} frequently outperform Reinforcement Learning (RL) baselines \cite{nisioti2025does}. 
By combining evolutionary search with local competition, algorithms such as MAP-Elites \cite{mouret2015illuminating} and Dominated Novelty Search (DNS) \cite{bahlous2025dominated} provide distinct advantages over standard Reinforcement Learning (RL) baselines in domains that explicitly require discovering a diverse repertoire of skills or transferring behaviours to new tasks \cite{nisioti2025does}.
However, this performance comes at a severe computational cost in sample efficiency. Performant QD methods often require tens of millions of environment interactions to deliver performant agents, limiting their practicality for expensive simulations and real-world robotics. 
Concurrently, the field of off-policy Deep RL has seen breakthroughs in sample efficient algorithms. High Update-to-Data (UTD) ratio algorithms, which perform more gradient updates per environment step, have allowed model-free algorithms to match the sample efficiency of model-based approaches \cite{chen2021randomized}. However, naive integration has been shown to not translate directly when applied to QD-RL algorithms \cite{lim2023understanding}. 
% {\color{blue} We should add a section here that discusses hybrid approaches (i.e. qdrl/pga-me stuff).} 

We posit that this failure stems partially from the inherent diversity of new trajectories generated by QD approaches. In traditional actor-critic RL, the optimal policy evolves slowly, allowing for a lagging target network (via Polyak averaging) to provide stable regression targets. In QD-RL, this target is derived from sampling the diverse population of agents maintained by QD, which expands rapidly and stochastically. A lagged target network fails to track this volatile and entropic frontier, introducing a misalignment between computed gradients and the true gradient derived from the current population, which is increasingly destructive with increased UTD. 

In this paper, we propose QDHUAC, a QD-RL algorithm that resolves the need for a target network via a target-free distributional critic architecture, allowing our critic to more directly track the diversity of the QD archive. To accommodate the high UTD ratios required for sample efficient learning, we utilise a novel hybrid normalisation architecture which constrains the critic's gradient magnitudes through combined weight and batch normalisation, creating a trust region in parameter space that remains stable under heavy distribution shifts of a diverse replay buffer. Our specific contributions are: 
\begin{enumerate*}
    \item We identify the Target Tracking Lag as the primary bottleneck for high UTD learning in evolutionary systems.
    \item We introduce QDHUAC, a comprehensive and sample-efficient Quality-Diversity Reinforcement Learning algorithm that enables stable, target-free learning at UTD ratios $\geq 10$.
    \item To achieve this stability within unstructured, non-stationary archives such as Dominated Novelty Search (DNS), we design a target-free, distributional, and residual critic architecture that utilises hybrid normalisation (weight and batch normalisation) to mitigate value divergence.
    \item We demonstrate that QDHUAC achieves comparable coverage and fitness on high-dimensional Brax locomotion tasks with an order of magnitude sample efficiency compared to QD-RL baselines.
\end{enumerate*}

\section{Background}
We define the Update-to-Data (UTD) ratio, denoted as $G$, as the number of gradient updates performed on the neural network parameters for every environment step collected. Standard online reinforcement learning algorithms typically operate in a low-UTD regime ($G\approx1$), tying the learning speed directly to the simulation throughput. In contrast, sample-efficient RL aims to maximise the information extracted from limited interactions by operating in a High-UTD regime ($G > 1$). While increasing G theoretically improves sample efficiency by re-using experience, in practice, high update ratios trained on off-policy rapidly lead to value overestimation and divergence \cite{d2022sample, lim2023understanding}.

\citet{lim2023understanding} demonstrated that standard QD-RL agents are significantly under-trained, and increased the update frequency to standard RL levels (UTD $\approx 1$) in order to increase returns. However, they also found that pushing beyond this to High-UTD regimes (e.g., using DroQ \cite{hiraoka2021dropout}) with UTD $=20$ yielded minimal performance increase at a large and impractical computational cost. We hypothesise that this failure to scale is due to the target network bottleneck. In architectures with target networks, performing iterations of updates against an outdated target network creates a feedback loop of stale value estimation, an effect which is amplified in the QD-RL setting. 

However, naively removing the target network severely degrades stability; \citet{d2022sample} identified the target network as the critical component allowing RL algorithms to scale with the replay ratio. To mitigate instability without target networks, prior work has relied on ensemble methods \cite{chen2021randomized, hiraoka2021dropout}, which incur significant computational costs, or periodic network resets \cite{d2022sample, schwarzer2023bigger}, which discard accumulated value information, leading to periodic performance collapses that disrupt the precise ranking required for evolutionary selection.

Recently, \citet{bhatt2019crossq} challenged the ubiquity of target networks, demonstrating that they are not a fundamental requirement for value estimation, but rather an architectural compensation for internal covariate shift \cite{ioffe2015batch}. By replacing the target network with Batch Normalisation, they proposed the CrossQ architecture, which allows the critic to track the current policy distribution instantaneously without the latency introduced by Polyak averaging. This 'zero-lag' property is particularly advantageous for the non-stationary frontiers of Quality-Diversity. However, the diverse and high-variance nature of QD archives exacerbates the risk of value divergence. Consequently, while CrossQ provides the theoretical foundation for synchronised learning, stabilising it within a population-based framework requires further regularisation to constrain the Lipschitz constant of the critic \cite{gogianu2021spectral}. 

\section{Target-Free Distributional Value Estimation}

Standard QD-RL methods rely on the actor-critic framework, where a critic $Q_{\theta}$ estimates the value of a policy $\pi_\phi$. This is a natural fit to a diverse population of actors, as a sound Q-function is valid across all policies. To ensure stability in Q learning, these methods universally employ a target network $Q_{\theta'}$, a lagging copy of the critic parameters updated via Polyak averaging $\theta' \leftarrow \tau\theta \; + \; (1\;-\;\tau)\theta'$. 
While effective for stationary RL tasks, in QD-RL, the optimal policy is not a single point but a rapidly expanding frontier of diverse behaviours. As the archive $\mathcal{A}$ discovers new diverse behaviours, the distribution of state-action pairs $(s, a)$ shifts drastically. While target networks effectively stabilise learning in stationary regimes, they inherently introduce a value estimation latency \cite{bhatt2019crossq}. In the context of QD, where the goal is rapid expansion of the behavioural frontier, this latency manifests as \textit{Update Staleness} \cite{fedus2020revisiting, tessler2019distributional}.

% A target network, by design, lags behind these shifts \cite{zhang2017deeper}. Consequently, when the population discovers a new and high-performing region of the descriptor space, the target network $Q_{\theta'}$ determines the value estimates from observations obtained from the previous frontier. This creates a target tracking lag, where the critic provides stale and pessimistic gradients that actively suppress the expansion of the archive. Removing the target network resolves this lag but typically leads to divergence due to unbounded value propagation. While traditional RL mitigates this with slow updates, the rapid frontier expansion in QD archives exacerbates this lag.

To enable stable target-free learning at high UTD ratios, we introduce a hybrid normalisation critic architecture. We decouple the stabilisation of the optimisation landscape from the stabilisation of the input distribution by combining Weight Normalisation (WN) \cite{salimans2016weight} and Batch Normalisation (BN) \cite{ioffe2015batch}. To prevent value divergence in adversarial settings such as actor-critic without a target network, we must constrain the gradient magnitude of the critic \cite{gulrajani2017improved} and apply weight normalisation to every affine layer $l$ in the critic:
\begin{equation}
    \textbf{W} = \frac{g}{||\textbf{v}||}\textbf{v}
\end{equation}
where $\textbf{v}$ is the parameter vector and $g$ is a learnable scalar. Unlike Spectral Normalisation \cite{gogianu2021spectral}, which strictly bounds the singular values, WN controls the gradient magnitude while preserving expressivity. 
% The replay buffer in QD-RL contains a mixture of diverse behaviours, which can create severe internal covariate shift. We apply BN after WN but before the activation function:
% \begin{equation}
%     y = \frac{h - \mu_\mathcal{B}}{\sqrt{\sigma^{2}_{\mathcal{B}} + \epsilon}} \cdot \gamma \; + \; \beta
% \end{equation}
% The normalisation of statistics across the minibatch $\mathcal{B}$ aids the robustness of the critic to the broad and diverse behaviours used to populate the replay buffer. 
The replay buffer in QD-RL contains a mixture of diverse behaviours, which can create severe internal covariate shift. We apply BN after WN but before the activation function:
\begin{equation}
    y = \frac{h - \mu_\mathcal{B}}{\sqrt{\sigma^{2}_{\mathcal{B}} + \epsilon}} \cdot \gamma \; + \; \beta
\end{equation}
where $h$ is the aggregated input signal from the preceding layer, $\mu_\mathcal{B}$ and $\sigma^{2}_{\mathcal{B}}$ are the mean and variance computed across the minibatch $\mathcal{B}$, $\epsilon$ is a small constant added for numerical stability, and $\gamma$ and $\beta$ are learnable scale and shift parameters. This normalisation of statistics across the minibatch $\mathcal{B}$ aids the robustness of the critic to the broad and diverse behaviours used to populate the replay buffer.

Furthermore, our use of BN serves to smooth the optimisation landscape \cite{santurkar2018does}. The combination of WN for optimisation stability and BN for distributional stability allows for high-UTD training without reducing the learning rate \cite{nikishin2022primacy}. To accurately estimate batch statistics for normalisation without a target network, we adopt the CrossQ technique of concatenating the current observations $s_t$ and next observations $s_{t+1}$ into a single batch. This ensures the batch normalisation layers estimate statistics over the full state distribution visited by the policies.

While CrossQ \cite{bhatt2019crossq} originally proposed a scalar critic with BN, we identify that scalar regression (MSE) is insufficient for the high-variance regime of QD. To address this, we employ a categorical distributional critic, as described in the C51 algorithm, \cite{bellemare2017distributional, dabney2018distributional}. In QD-RL, the replay buffer $\mathcal{D}$ contains transitions from thousands of distinct policies (the evolving population). To the critic, this diversity manifests as high aleatoric uncertainty as the same state $s$ may yield vastly different returns depending on which agent generated the transition. A scalar critic $Q(s,a)$ averages these returns into a single mean, discarding the underlying structure. In contrast, a distributional critic $Z(s,a)$ learns the full probability mass function, allowing for more general and accurate value estimation \cite{bellemare2017distributional}. Furthermore, scalar MSE loss provides sparse gradients, particularly when values plateau, and the cross-entropy loss used in distributional learning provides richer and denser gradient signals that accelerate representation learning in the early stages of training, a property crucial for sample efficiency \cite{hessel2018rainbow, clark2024beyond}. Additionally, in algorithms without a target network, scalar updates can cause large and oscillating updates that destabilise the actor. By distributing probability mass across fixed atoms, the distributional update is less volatile as mass shifts gradually between adjacent atoms rather than causing sudden spikes in the mean value \cite{lyle2019comparative}. 

Formally, we model the state-action value distribution $Z^{\pi}(s, a)$ using a discrete support of $N$ atoms $\{ z_i \}^N_{i=1}$, uniformly spaced between $V_{min}$ and $V_{max}$. The critic outputs a probability distribution $p_\theta(s, a)$ such that $Z^{\pi}(s, a) = \sum_i p_i(s, a)z_i$. The target distribution is computed via the projected Bellman update $\Phi \hat{\mathrm{T}} Z_\theta$: 
\begin{equation}
    \hat{\mathrm{T}} z_j = r + \gamma z_j, \quad \text{projected onto support} \; \{z_i\}
\end{equation}
where $\hat{\mathrm{T}}$ is the Bellman operator that applies the observed reward $r$ and discount factor $\gamma$ to shift the atoms, and $\Phi$ is the projection operator that maps the probabilities of these shifted atoms back onto the fixed original support $\{z_i\}$.
We minimise the Cross-Entropy loss between the current distribution and the projected target distribution, computed using the current critic:
\begin{equation}
    \mathcal{L}_{critic}(\theta) = \sum^N_{i=1} \; m_i \; \text{log} \, p_i(s, a; \theta)
\end{equation}
where $m_i$ is the projection of the current target distribution onto atom $i$. To ensure the return distribution fits within the support of the distributional critic $[V_{min}, V_{max}]$, we scale all environmental rewards by a factor of $0.01$.

With a robust and target-free critic $Z_\theta$, we can now optimise the policy $\pi_\phi$. Unlike standard deterministic gradients (TD3) \cite{fujimoto2018addressing}, we employ entropy regularisation \cite{haarnoja2018soft} to prevent the policy from collapsing into a single mode, a critical requirement for maintaining the diversity required by the QD archive \cite{eysenbach2018diversity}. The policy is updated to maximise the expected value of the distributional critic while maximising entropy $\mathcal{H}(\pi(\cdot|s))$: 
\begin{equation}
    J(\phi) = \mathbb{E}_{s \sim \mathcal{D}, a \sim \pi_\theta} \left[ \mathbb{E} \left[ Z^{\pi}(s, a) \right] + \alpha \mathcal{H}(\cdot | s) \right]
\end{equation}
where $\mathbb{E} \left[ Z^{\pi}(s, a) \right]$ is the mean value of the categorical distribution. The entropy temperature $\alpha$ is automatically tuned to maintain a target entropy, ensuring that the gradient-based learner remains stochastic enough to discover novel niches that the evolutionary operators might miss. Specifically, we treat $\alpha$ as a Lagrange multiplier optimised against the dual objective constraint $\bar{\mathcal{H}}$:

\begin{equation}
    J(\alpha) = \mathbb{E}_{\alpha \sim \pi_i} \left[ -\alpha \, \text{log} \, \pi_t \left(a|s_t \right) - \alpha\bar{\mathcal{H}} \right]
\end{equation}

where $\bar{\mathcal{H}}$ is the target entropy, set to $-|\mathcal{U}|$, the dimensionality of the action space $\mathcal{U}$, following standard practice \cite{haarnoja2018soft}.

Approximating the value distribution $Z^{\pi}$ across the heterogeneous and non-stationary support of a QD archive necessitates a function approximator with high representational capacity. However, scaling the depth of standard multi-layer perceptrons in RL frequently leads to optimisation instability due to gradient degradation in lower layers \cite{sinha2020d2rl, clark2024beyond, balduzzi2017shattered}. To reconcile the requirement for high capacity with the need for stable optimisation, we utilise a compact residual critic. This architecture utilises identity mappings to facilitate signal propagation during backpropagation. Unlike varying the depth in standard feed-forward networks, where gradients are strictly multiplicative through non-linearities, our architecture defines the output of a block $h_{l+1}$ as the summation of the input $h_l$ and a residual transformation $\mathcal{F}$ \cite{he2016identity}: 
\begin{equation} h_{l+1} = \underbrace{h_l}_{\text{Identity Mapping}} + \underbrace{\mathcal{W}_2 \left( \sigma \left( \mathcal{W}1 (h_l) \right) \right)}_{\text{Residual Mapping } \mathcal{F}(h_l)} \end{equation}
where $\mathcal{W}$ denotes the hybrid (WN + BN) normalised linear layers. The identity mapping preserves the magnitude of the gradient signal, mitigating vanishing gradients in high-UTD training regimes and ensuring that the Bellman error effectively updates the feature extraction layers. This architecture for the critic was specifically introduced by \citet{nauman2024bigger}, which demonstrated that while standard RL architectures saturate at small parameter counts, combining deep residual connections with strong regularisation allows the critic to maintain plasticity at high UTD ratios. 

We adapt the 'Optimistic' component from \citet{nauman2024bigger} to the distributional setting. In the original scalar implementation, optimism is enforced by explicitly initialising the final layer bias to output $V_{max}$, forcing the agent to explore by overestimating values at initialisation. In contrast, our approach achieves this optimism implicitly through the maximum entropy property of the distributional head. At initialisation, standard weight distributions yield output logits near zero, resulting in a uniform probability mass over the atoms $\{z_i\}$. Consequently, the initial value estimate defaults to the mean of the support $\mathbb{E}[Z] \approx \frac{V_{min} + V_{max}}{2}$. By setting $V_{max}$ to a realistic upper bound on returns, this uniform prior effectively acts as a maximum entropy regulariser on the value estimates, encouraging the agent to explore state-action pairs where the critic has not yet concentrated probability mass.

To manage the population of diverse strategies, we employ Dominated Novelty Search (DNS) \cite{bahlous2025dominated} rather than the structured archive typically used in MAP-Elites. While MAP-Elites relies on a rigid, pre-defined tessellation of the descriptor space, which imposes artificial discretisation artifacts \cite{vassiliades2017using} and necessitates careful tuning of grid resolution \cite{janmohamed2025multi}, DNS maintains an unstructured population (or 'repertoire') of fixed size. Survivors are selected based on 'dominated novelty', a meta-fitness defined as the average distance in descriptor space to the $k$-nearest neighbours that possess superior fitness. This metric creates a selection pressure that naturally identifies and maintains a balance between performance versus diversity. DNS allows the population to adapt dynamically to the rapidly shifting geometry of the solution space, a property that is critical when coupled with a High-UTD learner that can swiftly alter the scale and distribution of reachable behaviours.

At each generation, the DNS archive is evolved via a hybrid mutation strategy. Firstly, we apply gene-level Iso+Line mutations \cite{vassiliades2017using} to ensure broad coverage of the descriptor space. Secondly, we perform purely gradient-based updates on a subset of parents to accelerate local exploitation. Finally, we inject the High-UTD gradient-optimised policy $\pi_\phi$. The gradient policy exploits the global value landscape learnt by the critic, while the gene-level mutations ensure broad coverage of the descriptor space. Collectively, these agents explore the search space broadly, unconstrained by the limitations of a single local optimisation. Transitions generated by this heterogeneous population are aggregated in $\mathcal{D}$, allowing the target-free critic to learn from a diverse distribution of behaviours. Consequently, when $\pi_\phi$ is re-injected, it benefits from High-UTD training on this diverse experience to aggressively exploit the most promising niches of the search space discovered by the population. To further improve sample efficiency, we employ Prioritised Experience Replay \cite{schaul2015prioritized}, sampling transitions proportional to their temporal difference error with a priority exponent $\omega=0.6$. We formalise the combination of these elements in Algorithm \ref{alg:qdhuac}.

 % Additionally, we apply purely gradient-based updates to a subset of the parents from the archive to accelerate local exploitation. 

\begin{algorithm}[tb]
\caption{QDHUAC (Quality-Diversity High Update Actor-Critic)}
\label{alg:qdhuac}
\begin{algorithmic}[1]
\STATE \textbf{Initialise:} Policy $\pi_\phi$, Distributional Critic $Z_\theta$, Archive $\mathcal{A}$, Replay Buffer $\mathcal{D}$
\STATE \textbf{Initialise:} Emitter $\mathcal{E}$ (e.g., PGA-ME with genetic operators)
\STATE \textbf{Hyperparameters:} Learning rates $\eta_{actor}, \eta_{critic}$, Temperature $\alpha$, UTD Ratio $U$
\FOR{generation $g = 1, \dots, G$}
    % \STATE \textbf{1. Evolutionary Rollout}
    \STATE Generate population of genotypes $\{ \theta_i \}_{i=1}^N \sim \mathcal{E}(\mathcal{A})$
    \STATE Evaluate $\{ \theta_i \}$ in environment to get fitness $f_i$, descriptor $d_i$, and transitions $\tau_i$
    \STATE Add $\{ \theta_i, f_i, d_i \}$ to Archive $\mathcal{A}$
    \STATE Add transitions $\tau_i$ to Replay Buffer $\mathcal{D}$
    
    % \STATE \textbf{2. Target-Free High-UTD Update}
    \FOR{$j = 1, \dots, N \times U$}
        \STATE Sample batch $B \sim \mathcal{D}$
        \STATE \textbf{Critic Update:} Compute distributional loss $\mathcal{L}_{critic}$ using $Z_\theta$
        \STATE $\theta \leftarrow \theta - \eta_{critic} \nabla_\theta \mathcal{L}_{critic}$
        
        \STATE \textbf{Actor Update:} Compute max-entropy loss $\mathcal{L}_{actor}$ 
        \STATE $\phi \leftarrow \phi - \eta_{actor} \nabla_\phi \mathcal{L}_{actor}$
        \STATE Update temperature $\alpha$ automatically (Eq. 5)
    \ENDFOR
    
    % \STATE \textbf{3. Injection}
    \STATE Inject RL actor $\pi_\phi$ into Archive $\mathcal{A}$
\ENDFOR
\end{algorithmic}
\end{algorithm}

\section{Empirical Methodology}
To evaluate the sample efficiency and stability of our proposed method, we conduct experiments on high-dimensional continuous control tasks. We implement our algorithm using the QDax framework \cite{chalumeau2024qdax}.

\subsection{Environments}
We select five challenging locomotion tasks from the standard QDax suite \cite{chalumeau2024qdax}. These environments are chosen for their high-dimensional state-action spaces and complex dynamics, providing a rigorous testbed for stability in high-UTD regimes.

\begin{itemize}
    \item \textbf{Hopper} ($|\mathcal{S}|=11, |\mathcal{A}|=3$): A monopedal robot that must hop forward while maintaining balance.
    \item \textbf{Walker2D} ($|\mathcal{S}|=17, |\mathcal{A}|=6$): A bipedal robot that must balance and hop forward.
    \item \textbf{HalfCheetah} ($|\mathcal{S}|=18, |\mathcal{A}|=6$): A fast two-legged robot requiring stable gait coordination.
    \item \textbf{Ant} ($|\mathcal{S}|=87, |\mathcal{A}|=8$): A quadruped with a large state space, requiring coordination of multiple legs.
    \item \textbf{Humanoid} ($|\mathcal{S}|=227, |\mathcal{A}|=17$): The most challenging high-dimensional control task, often prone to early termination (falling) which destabilises standard RL critics.
\end{itemize}

\subsection{Baselines}
To validate the contributions of our target-free distributional architecture, we compare against established Quality-Diversity baselines in PGA-ME and QD-PG. PGA-ME combines genetic variation with a TD3-based actor-critic. Crucially, PGA-ME relies on target networks and typically operates at a low-UTD ratio. Comparing against PGA-ME isolates the benefits of our high-UTD, target-free critic. Quality-Diversity Policy Gradient (QD-PG), a fully gradient-driven approach, utilises policy gradients for both quality and diversity maximisation via dual critics. This comparison serves as a reference for a state-of-the-art QD-RL algorithm that fully utilises differentiable diversity objectives but remains constrained by standard target network updates. We use the hyperparameters from the original implementations \cite{nilsson2021policy, pierrot2022diversity}, with the exception of an environment length of $1000$ steps and an environment batch size of $10$.

% \subsection{Performance Metrics}
We evaluate performance using standard Quality-Diversity metrics \cite{pugh2016quality}. Firstly, QD-Score, defined as the sum of fitnesses of all solutions in the archive. This metric captures both the diversity (archive size) and the quality (fitness) of the population. Secondly, coverage, defined as the proportion of the descriptor space discovered by the algorithm. Thirdly, to evaluate sample efficiency, we measure the Area Under the Curve (AUC) of the QD-Score over the training horizon. This penalises methods that eventually converge but require significantly more samples to do so. Finally, we measure the max fitness of the archive.

% \subsection{Implementation Details \& Hyperparameters}

We conduct our experiments using the JAX/MJX backend to enable massive parallelisation \cite{freeman2021brax}. We employ a C51 distributional critic with $N=101$ atoms support over the range $[V_{min}, V_{max}]$. To accommodate the varying reward scales of different environments, we apply a reward scaling factor of $0.01$ to fit returns within the critic's support. The full list of hyperparameters and architectural details used for QDHUAC can be found in Appendix \ref{app:hyperparameters}.

% For the genetic variation step, we utilise the Iso+Line mutation operator \cite{vassiliades2018discovering} with $\sigma_{iso}=0.005$ and $\sigma_{line}=0.1$. Additionally, for the gradient-based mutation operator, we perform $10$ steps of Adam optimisation with a learning rate of $5e^{-3}$ on the selected parents before adding them to the offspring batch.

% \textit{Training Regime:} {\color{red} We train for $2000$ generations. At each generation, we perform $N_{grad}=10,000$ gradient updates against a replay buffer of size $10^6$. This corresponds to a Replay Ratio (RR) of approx. $25$ updates per environment step. The critic and actor are optimised using AdamW with learning rates of $3 \cdot 10^{-4}$. To further enhance sample efficiency, we sample from the replay buffer using Prioritised Experience Replay (PER) with a priority exponent of $0.6$.}

\subsection{Results}
\begin{figure*}[h]
    \centering
    \includegraphics[width=0.85\linewidth]{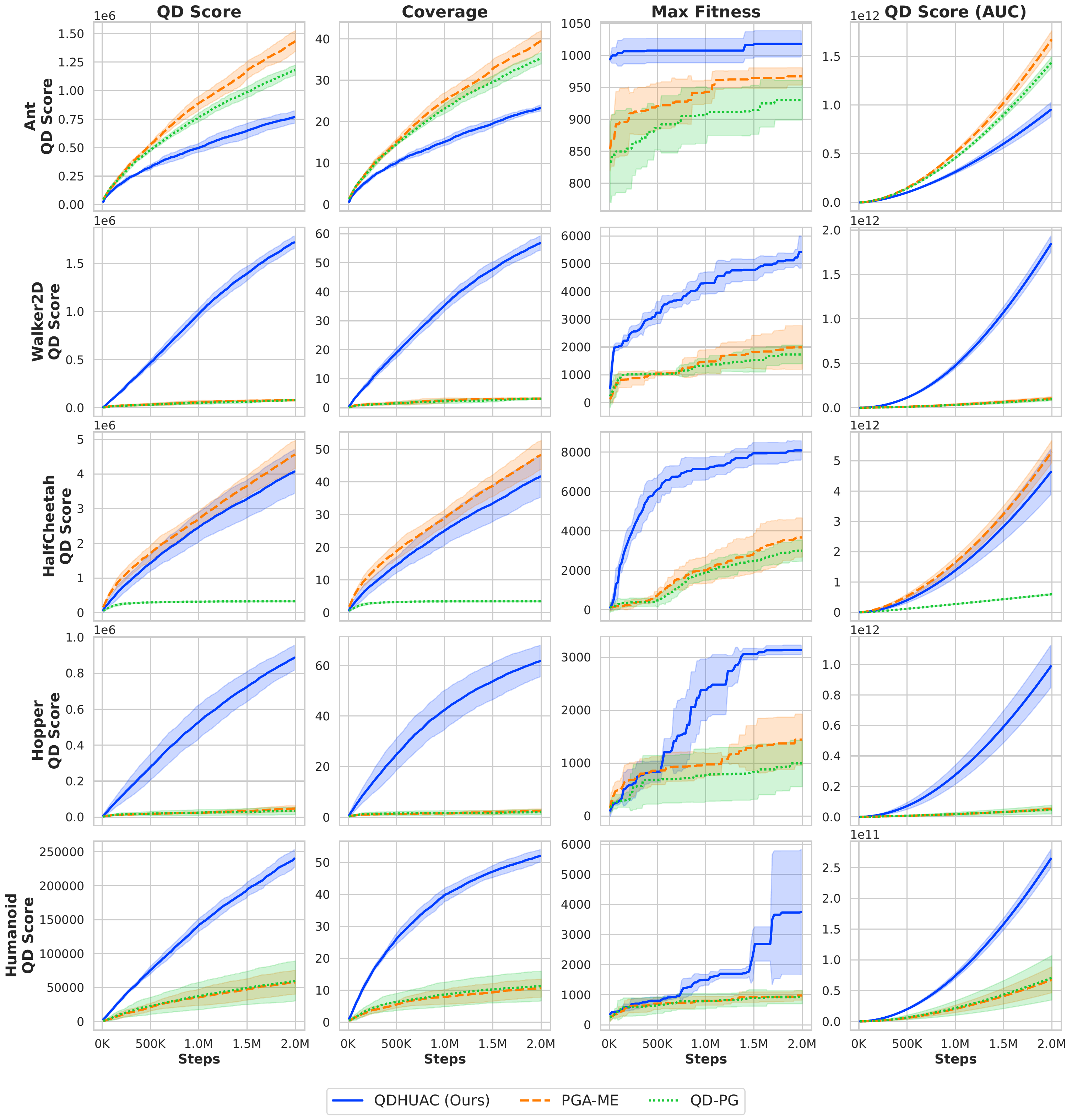}
    \caption{\textit{Benchmarking Sample Efficiency and Performance.} We compare QDHUAC (Ours) against baselines across five continuous control locomotion tasks. QDHUAC demonstrates improvements in sample efficiency, achieving higher scores earlier than baselines in complex environments such as Humanoid and Ant. Coverage is measured as \% of projected archive covered, as seen in \citet{bahlous2025dominated}.}
    \label{fig:main_results}
\end{figure*}

To validate the efficacy of our target-free distributional framework, we benchmark QDHUAC against PGA-ME \cite{lim2023understanding, nilsson2021policy} and QD-PG \cite{pierrot2022diversity}, two QD-RL based approaches. We evaluate performance across five standard continuous control locomotion tasks from the Brax suite via QDax \cite{freeman2021brax, chalumeau2024qdax}. All methods were trained for 2 million environment steps, with an environment batch size of 10. To accurately compare between algorithms using structured archives such as PGA-ME and QD-PG, and unstructured archives such as QDHUAC, we evaluate coverage using a unified passive archive. Specifically, all behaviours discovered by all algorithms during training are projected onto a standard, pre-defined MAP-Elites grid solely for the purpose of calculating the coverage metric, ensuring that the structural differences in their native archives do not skew the evaluation.

% Figure \ref{fig:main_results} provides the results comparing the performance of QDHUAC against the PGA-ME and QD-PG baselines across the benchmarked environments for $N=5$ random seeds, illustrating the mean and standard deviation for QD Score, Coverage, Max Fitness, and QD Score (AUC). 

The empirical results presented in Figure \ref{fig:main_results} validate our core hypothesis: that the primary bottleneck in applying sample-efficient RL to Quality-Diversity is not flawed integration strategies, but the Target Tracking Lag introduced by standard target networks. By removing this bottleneck, QDHUAC demonstrates that High-UTD learning can be effectively stabilised in population-based search, leading to superior optimisation and coverage in complex exploration domains, while returning increases in maximum fitness across all evaluated tasks.

In environments with difficult exploration dynamics, such as Walker2D and Hopper, the baseline methods fail to achieve significant QD scores. In these tasks, the current "optimal" policy is not a static target but a rapidly expanding frontier. Standard target networks, which update via Polyak averaging, lag behind this frontier. This latency causes the critic to misidentify novel, high-performing solutions as low-value (False Negatives), effectively suppressing the discovery of new niches. In contrast, QDHUAC’s target-free critic more closely tracks the population's true value distribution, encouraging the exploitation of new niches as they are discovered. 

The rapid initial rise in QD Score across all environments suggests that the Distributional Critic (C51) provides superior learning signals compared to the scalar critics used in PGA-ME. In the high-variance regime of a diverse replay buffer, a scalar Mean Squared Error (MSE) loss often washes out the underlying structure of returns. The cross-entropy loss used in QDHUAC provides richer, denser gradient signals that accelerate representation learning in the early stages. This is particularly advantageous in Ant, where QDHUAC achieves high fitness early in training, leveraging the distributional head to navigate the complex morphology despite the high aleatoric uncertainty of the archive.

% While QDHUAC achieves high Max Fitness in the Ant environment, we observe a reduction in total coverage compared to PGA-ME. We hypothesise that this is a byproduct of the aggressive exploitation pressure introduced by High-UTD gradient updates. In high-dimensional control tasks like Ant, the gradient signal strongly guides the policy toward stable, high-performing gaits. Since QDHUAC updates the policy magnitude more frequently than the baseline, the population converges rapidly toward these high-fitness regions, potentially at the expense of exploring lower-fitness regions of the descriptor space. However, the significantly higher Max Fitness suggests that this trade-off is beneficial; QDHUAC may cover less total area, but it prioritises populating the archive with elite, high-quality solutions rather than spreading into low-performing regions often traversed by purely divergent search.
While QDHUAC achieves significantly higher Max Fitness in the Ant and HalfCheetah environments, we observe a reduction in total coverage compared to PGA-ME. We hypothesise that this is a byproduct of the aggressive exploitation pressure introduced by High-UTD gradient updates. In high-dimensional control tasks, the gradient signal strongly guides the policy toward stable, high-performing gaits. Since QDHUAC updates the policy magnitude more frequently than the baseline, the population converges rapidly toward these high-fitness regions, potentially at the expense of exploring lower-fitness regions of the descriptor space. However, the massive gains in Max Fitness (e.g., reaching nearly 8000 in HalfCheetah compared to the baseline's 3500) suggest that this trade-off is highly beneficial; QDHUAC prioritises populating the archive with elite, high-quality solutions rather than spreading into low-performing regions often traversed by purely divergent search.

% In the Humanoid task, QDHUAC achieves a fitness of $\approx 5000$ within the first evaluation update. This corresponds to the agent learning a policy which survives for the full length of the environment within the very first interaction batch. Standard baselines perform significantly fewer updates per iteration, insufficient to find a policy which maximises survival time and therefore, reward. In contrast, the increased UTD allows the QDHUAC agent to find a policy which survives for the length of the episode and collect a greater reward.
In the Humanoid task, which presents the most challenging high-dimensional state and action space, QDHUAC demonstrates increased sample efficiency in discovering diverse behaviours. While its maximum fitness climbs in a step-wise manner and surpasses the baselines to reach nearly 4000, its ability to rapidly construct a diverse repertoire of distinct and surviving gaits is demonstrated in the QD-Score. We hypothesise that the increased UTD allows the population to find surviving gaits earlier, in contrast to the standard baselines, which struggle to find these behaviours.
 
In summary, QDHUAC outperforms baselines by resolving the structural conflict between High-UTD learning and QD. By discarding the target network, we eliminate the lag that blinds standard critics to rapid evolutionary progress, where the hybrid normalized architecture provides the stability required to exploit these gains. 
% QDHUAC demonstrates strong performance across the majority of the tested environments, consistently outperforming baseline in max fitness, indicating a superior capability for finding high-performing solutions. In particular, QDHUAC presents a significant performance advantage in the complex and high-dimensional Ant and Humanoid environments within the limited number of available environment steps. In the Walker2D environment, the failure of the baselines to progress beyond a Max Fitness of approximately 1000 suggests the presence of deceptive or local optima or difficult exploration dynamics that 

% \subsubsection{{\color{red} DNS vs Map-Elites Ablation}}

% \subsubsection{Analysis of Value Estimation Latency}

\begin{figure*}[t]
    \centering
    \begin{minipage}{0.48\textwidth}
        \centering
    \includegraphics[width=0.9\linewidth]{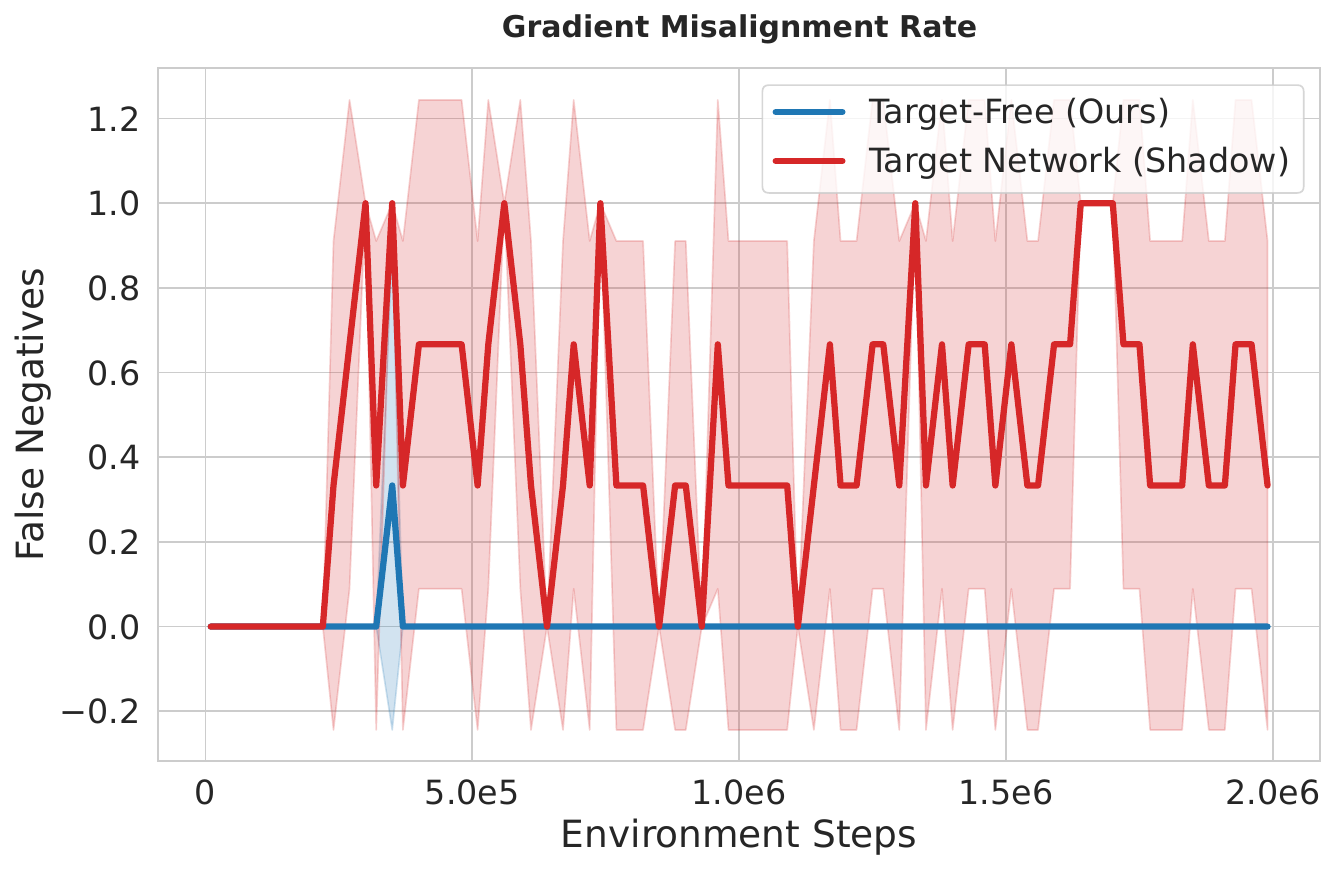}
    \caption{\textit{Deconstructing the Pessimism Gap in Target Networks.} To quantify the impact of target networks on discovery of diverse behaviours, we trained a standard Polyak-averaged target network (labelled "Shadow Target", with $\tau=0.005$) alongside our target-free critic on HalfCheetah ($N=5$). The shadow target underestimated the true discounted return during periods of rapid policy improvement, creating a pessimism gap. This underestimation leads to a high FNR, where the target network misidentifies improving agents as low-value, suppressing the development of niches of new behaviours.}
    \label{fig:false_negative_rate}
    \end{minipage}\hfill
    \begin{minipage}{0.48\textwidth}
    \centering
    \includegraphics[width=0.9\linewidth]{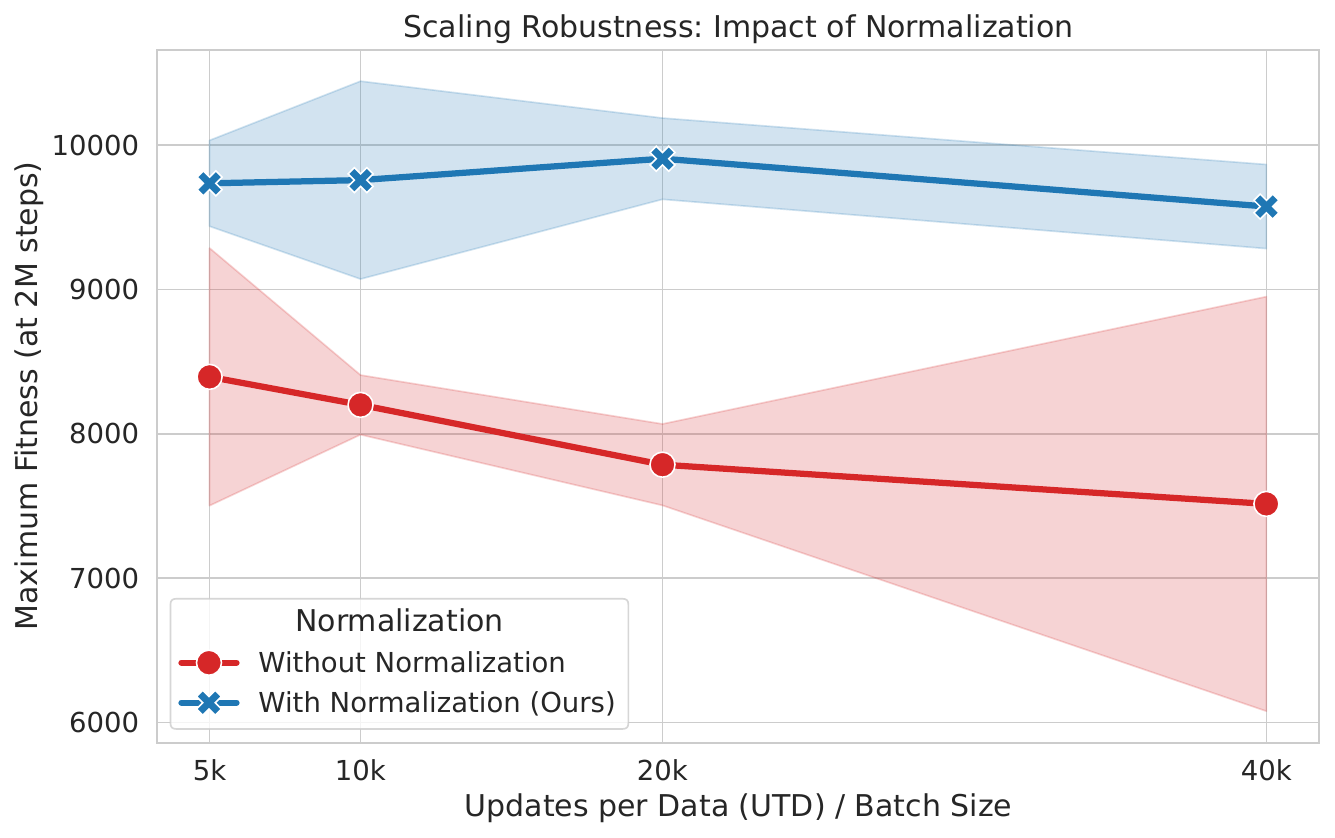}
    \caption{\textit{Scaling Robustness and the Role of Hybrid Normalisation.} We evaluate the stability of our critic under increasing Update-to-Data (UTD) ratios, ranging from $5k$ to $40k$ updates per iteration on HalfCheetah ($N=5$). While the standard unnormalised critic suffers from performance degradation at high UTD ratios, losing $\approx 30\%$ of its peak fitness as updates increase, the batch and weight normalised layers significantly mitigate this instability. The normalised critic maintains high performance across the sweep, with a performance peak at 20,000 updates per iteration ($UTD = 2$), demonstrating that distributional stability is required for high-UTD training.}
    \label{fig:scaling_fitness_utd}
    \end{minipage}
    \vspace{-0.5cm}
\end{figure*}

% \begin{figure}
%     \centering
%     \includegraphics[width=0.9\linewidth]{assets/gradient_misalignment.pdf}
%     \caption{\textit{Deconstructing the Pessimism Gap in Target Networks.} To quantify the impact of target networks on discovery of diverse behaviours, we trained a standard Polyak-averaged target network (labelled "Shadow Target", with $\tau=0.005$) alongside our target-free critic on HalfCheetah ($N=5$). The shadow target underestimated the true discounted return during periods of rapid policy improvement, creating a pessimism gap. This underestimation leads to a high FNR, where the target network misidentifies improving agents as low-value, suppressing the development of niches of new behaviours.}
%     \label{fig:false_negative_rate}
% \end{figure}

\textit{Analysis of Value Estimation Latency.} Standard QD-RL algorithms rely on Polyak-averaged target networks to stabilise value estimation. We hypothesised that in the highly non-stationary regime of Quality-Diversity search, this averaging introduces a \textit{Target Tracking Lag} that creates a pessimistic bias, causing the rejection of novel, high-performing solutions

While QDHUAC trains using a Target-Free critic $Q_\theta$, we simultaneously maintained a target network $Q_{\theta'}$. This network was not used for gradient updates but was updated at every step using standard Polyak averaging with $\tau=0.005$

% :
% \begin{equation} 
% \theta' \leftarrow \tau\theta + (1-\tau)\theta' 
% \end{equation}

By logging the predictions of $Q_{\theta'}$ alongside $Q_{\theta}$, we measure the precise opportunity cost and latency introduced by the target network mechanism in empirical environments. We computed the actual discounted return $G_t$ for the top performing agents in each batch and raw environmental rewards were unscaled by the reward scaling factor $\lambda^{-1}$ prior to discounting: 
\begin{equation} G_t = \sum_{k=0}^{T} \gamma^k (\lambda^{-1} r_{t+k}) \end{equation}
This provides an objective measure of the true value of the current policy, independent of the critic's estimation. We define the pessimism gap $\Delta_Q$ as the difference between the target free estimate and the target network estimate:
% \begin{equation} 
$\Delta_Q = \mathbb{E}_{s \sim \mathcal{D}}[Q_{\theta}(s, \pi(s))] - \mathbb{E}_{s \sim \mathcal{D}}[Q{_\theta'}(s, \pi(s))]$ 
% \end{equation}
A positive gap $(\Delta_Q > 0)$ indicates that $Q_{\theta'}$ is estimating higher values than the target network (optimism), while a negative gap indicates the reverse. 
Furthermore, we measure the \textit{False Negative Rate} (FNR), defined as the frequency with which a solution achieves high fitness ($G_t > \eta_{90}$) but the critic estimates a value lower than the current archive threshold ($Q < \eta_{90}$). 
% \begin{equation} 
$\text{FNR} = \frac{\sum \mathbb{I}(G_t > \eta_{90} \land \hat{Q} < \eta_{90})}{\sum \mathbb{I}(G_t > \eta_{90})} $
% \end{equation}
A high FNR indicates that the critic is providing pessimistic gradient updates that actively discourage the policy from exploiting valid, high-performing regions of the descriptor space. While the archive itself uses environmental fitness for retention, the policy optimiser relies entirely on the critic. A false negative here represents a critical failure in the learning signal; the critic tells the actor that a high-value region is poor, effectively suppressing the gradient-based discovery of that niche. 

Furthermore, the critic in our architecture governs generation, not selection. The selection mechanism (DNS) is grounded in reality; it uses the actual, ground-truth fitness returned by the environment to determine whether a new child is integrated into the population. The Q-network, however, drives the gradient-based mutation operator, guiding the policy update $\Delta_{\phi}Q(s, \pi_{\phi}(s))$ towards regions of ostensibly higher value. Therefore, when we analyse the impact of the critic on discovery, we refer to the quality of the offspring proposed by the gradient update. If the critic is pessimistic or inaccurate (as seen with the shadow target network), the gradient ascent step directs the policy toward sub-optimal manifolds. Despite being optimised against the critic, these generated children possess low true fitness and are therefore eliminated by the rigorous selection filter of DNS. Consequently, a poor Q-network does not mis-rank the archive but actively destabilises the generative process, producing a stream of non-viable candidates that fail to survive, thereby stalling the expansion of the archive.

Figure \ref{fig:false_negative_rate} provides the FNR for $N=5$ independent seeds in HalfCheetah. We observe that the standard target network systematically fails to track rapid fitness improvements, resulting in a persistent FNR that acts as a 'pessimism penalty' on the learning process. Conversely, our target-free architecture achieves negligible misalignment, ensuring that the policy receives accurate, optimistic gradients. This suggests that the latency introduced by target networks is not merely a theoretical concern, but a structural bottleneck that actively prevents the policy from consolidating the gains made by the evolutionary search.

% \subsubsection{High-UTD vs Low-UTD Ablation}
% \begin{figure}[h]
%     \centering
%     \includegraphics[width=0.9\linewidth]{assets/ablation_scaling_fitness.pdf}
%     \caption{\textit{Scaling Robustness and the Role of Hybrid Normalisation.} We evaluate the stability of our critic under increasing Update-to-Data (UTD) ratios, ranging from $5k$ to $40k$ updates per iteration on HalfCheetah ($N=5$). While the standard unnormalised critic suffers from performance degradation at high UTD ratios, losing $\approx 30\%$ of its peak fitness as updates increase, the batch and weight normalised layers significantly mitigate this instability. The normalised critic maintains high performance across the sweep, with a performance peak at 20,000 updates per iteration ($UTD = 2$), demonstrating that distributional stability is required for high-UTD training.}
%     \label{fig:scaling_fitness_utd}
% \end{figure}

\textit{High-UTD vs Low-UTD Ablation.} We hypothesise that removal of the target network enables higher sample efficiency, provided that the critic can remain stable. To validate the necessity of BN and WN in our network architecture, we evaluated the agent's performance across a sweep of Update-to-Data (UTD) ratios, ranging from 5,000 (UTD=0.5) to 40,000 (UTD=4) updates per iteration. Figure \ref{fig:scaling_fitness_utd} illustrates the impact of normalisation on scaling robustness on the HalfCheetah environment with $N=5$ unique seeds. We observe that the standard unnormalised distributional critic struggles to utilise the higher UTD and suffers from performance degradation as the UTD ratio increases further. At 40k updates per iteration, the unnormalised agent loses approximately $30\%$ of its maximum fitness capacity, indicating that the critic begins to overfit to the replay buffer, losing its ability to generalise to new transitions.

In contrast, the hybrid normalisation architecture demonstrates greater stability, effectively mitigating divergence and allowing the agent to maintain high performance at higher update frequencies, a core requirement for environment step constrained algorithms. Notably, QDHUAC achieves its peak performance at 20,000 updates per iteration, balancing sample efficiency with training stability. While there is a minor saturation effect observed at 40k updates per iteration, common in high-UTD methods such as REDQ \cite{chen2021randomized}, the normalised critic remains more robust than the baseline, confirming that architectural regularisation is a requirement for target-free learning in high-update QD-RL. 

% \subsubsection{Isolating the Impact of Hybrid Normalisation}

\begin{figure}
    \centering
    \includegraphics[width=\linewidth]{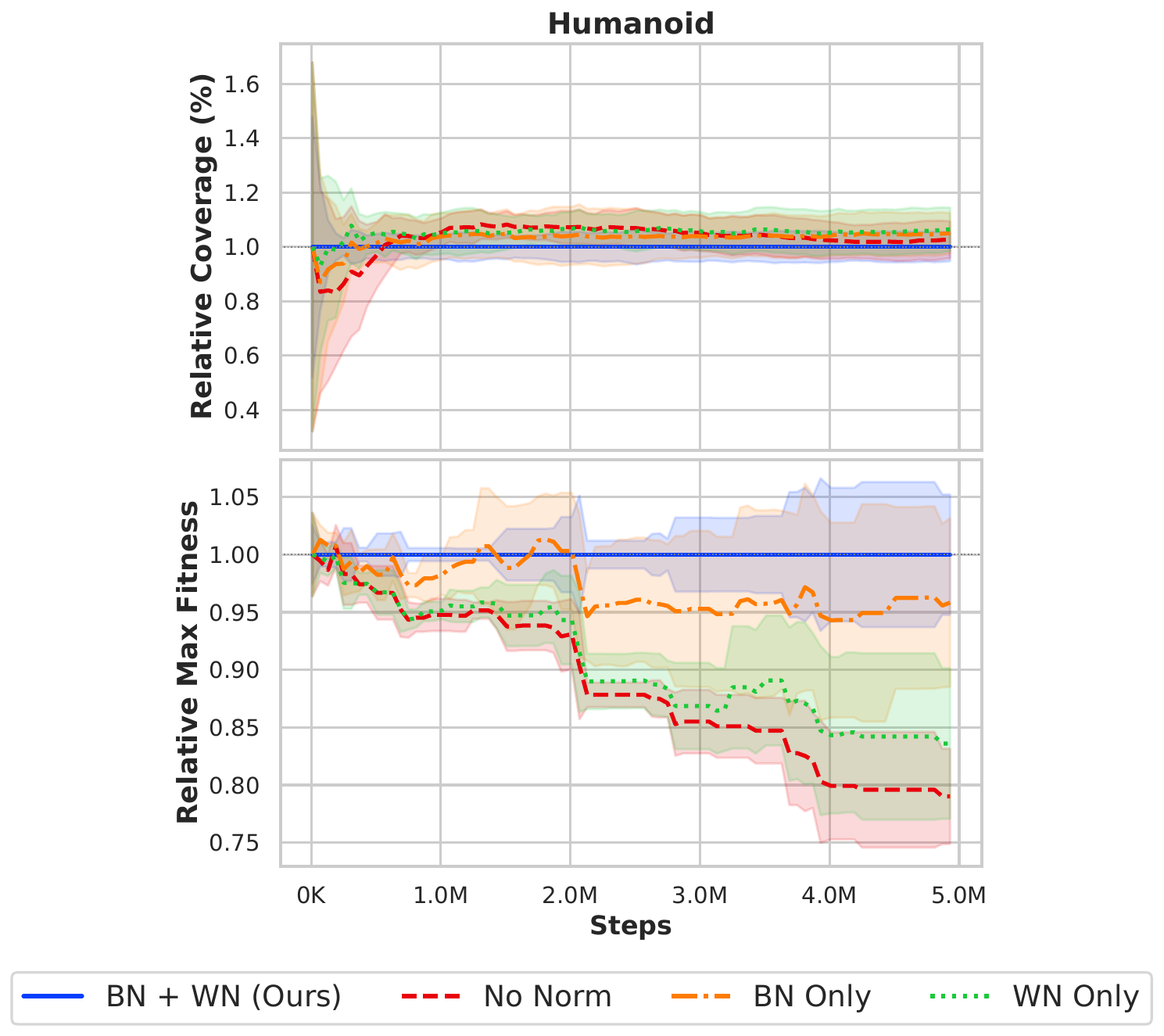}
    \caption{\textit{Ablation of Critic Normalisation Components.} Relative performance of normalisation schemes on the Humanoid environment, normalised against the full QDHUAC architecture (BN + WN, solid blue line). While coverage is minimally affected, combining both Batch and Weight Normalisation is uniquely required to prevent long-term value divergence and maintain peak maximum fitness.}
    \label{fig:norm_types}
    \vspace{-0.5cm}
\end{figure}

\textit{Isolating the Impact of Hybrid Normalisation.} To address whether both Weight Normalisation (WN) and Batch Normalisation (BN) are strictly necessary, we isolated their effects on the Humanoid environment. Figure \ref{fig:norm_types} shows the choice of normalisation has a negligible effect on overall descriptor Coverage. However, it has a profound impact on Maximum Fitness. We observe that an unnormalised critic ("No Norm") and a critic using only Weight Normalisation ("WN Only") suffer severe performance degradation, losing up to 20\% of their fitness capacity. While using Batch Normalisation alone ("BN Only") initially stabilises the network, supporting findings from CrossQ \cite{bhatt2019crossq}, it begins to lose plasticity and degrade after 2 million steps. The combination of both ("BN + WN") maintains stable, peak exploitation performance throughout the entire training run. This suggests that while BN is sufficient to anchor the shifting input distribution of the archive, WN is required to strictly bound the Lipschitz constant and prevent late-stage gradient divergence under high-UTD regimes.

Finally, to ensure that our sample efficiency gains are attributable to the target-free critic rather than the specific choice of evolutionary archive, we conducted an ablation study substituting Dominated Novelty Search (DNS) with a standard MAP-Elites grid. Details and results of this ablation can be found in Appendix \ref{app:me_abl}.

\section{Related Work}
The integration of a gradient-based optimisation into evolutionary search has been a primary driver of performance in continuous control QD. \citet{nilsson2021policy} and \citet{pierrot2022diversity} demonstrated that integrating off-policy gradient updates alongside evolutionary variation significantly accelerates the discovery of high-performing policies. However, these methods inherit the limitations of standard actor-critic architectures; they rely on target networks to stabilise learning, which restricts them to low UTD ratios to avoid divergence, a restriction that is incompatible with low data QD-RL or sample efficient QD-RL. While \citet{lim2023understanding} explored increasing the UTD ratio in QD-RL, they found that performance plateaued or degraded due to the instability of off-policy updates on diverse data. While recent works have attempted to scale these approaches, such as \citet{xue2024sample}, which improved efficiency through architectural decomposition, they do not address the fundamental sample-inefficiency of the critic update itself, leaving the potential of high-UTD learning untapped.

Recent advances in sample-efficient RL has focused on increasing the UTD ratio. Algorithms such as REDQ \cite{chen2021randomized} and DroQ \cite{hiraoka2021dropout} utilise ensemble methods to mitigate the bias-variance tradeoff at high-UTD ratios, but at a significant computational cost. Recent work has identified the "primacy bias" \cite{nikishin2022primacy}, where agents overfit to early data, while \citet{ma2023revisiting} demonstrated that high replay ratios lead to a loss of plasticity in the value network. To combat this, approaches such as SPEQ \cite{romeo2025speq} propose offline stabilisation phases and \citet{hussing2024dissecting} highlighted the necessity of combatting value divergence through aggressive regularisation. CrossQ \cite{bhatt2019crossq} proposed replacing the target network with Batch Normalisation \cite{ioffe2015batch} to enable target-free learning. While QDHUAC shares the target-free approach in \citet{bhatt2019crossq}, QDHUAC extends this direction through a distributional target-free critic with hybrid normalisation, which we demonstrate is critical for maintaining stability when learning from the non-stationary distributions of a QD archive.

Distributional RL \cite{bellemare2017distributional, dabney2018distributional, lyle2019comparative, bellemare2023distributional} models the full distribution of returns rather than just the mean, and improves scalability of networks, even in problems which are naturally framed as regression problems \cite{farebrother2024stop}. The C51 algorithm \cite{bellemare2017distributional} demonstrated that categorical distributions provide richer gradient signals and improved stability in discrete settings. QDHUAC accelerates representation learning in the early stages of evolution by utilising the dense gradients of the cross-entropy loss, an property essential for escaping the local optima found in high-dimensional locomotion tasks.

\vspace{-0.1cm}
\section{Conclusion}
In this work, we identified the Target Tracking Lag as a fundamental structural bottleneck preventing the application of sample-efficient Reinforcement Learning to Quality-Diversity optimisation. We demonstrated that standard target networks, while stabilising stationary RL tasks, fail to track the rapidly expanding frontier of an evolutionary population, resulting in a pessimistic value bias that suppresses discovery.

To resolve this, we introduced QDHUAC, a novel QD-RL algorithm that leverages a Target-Free Distributional Critic. By combining Hybrid Normalisation (Weight and Batch Normalisation) with the dense gradient signals of a categorical distribution, QDHUAC enables stable learning at higher Update-to-Data (UTD) ratios without the latency of Polyak averaging. Our empirical results on high-dimensional locomotion tasks confirm that this architecture significantly improves sample efficiency, achieving high Max Fitness scores with an order of magnitude fewer environment steps than existing baselines. These findings suggest that the future of efficient evolutionary learning lies not merely in scaling update ratios, but in designing critic architectures that are robust to the non-stationarity inherent in population-based search.

\clearpage
\bibliographystyle{ACM-Reference-Format}
\bibliography{acmart}
\clearpage
\section*{Appendix}
\subsection{Hyperparameters for QDHUAC} \label{app:hyperparameters}
\begin{table}[h]
    \centering
    \caption{\textit{Hyperparameters for QDHUAC. We separate parameters into three categories: general training settings (shared across environments), the specific architecture and support range for the Distributional Target-Free Critic, and the evolutionary parameters for Dominated Novelty Search (DNS). All environments use the same hyperparameters, except for the support range, which are adjusted to match the return scale of each task. }}
    \label{tab:hyperparameters}
    \begin{tabular}{l|l}
        \toprule
        \textbf{Parameter} & \textbf{Value} \\
        \midrule
        \multicolumn{2}{c}{\textit{General Training}} \\
        \midrule
        Optimiser & Adam \\
        Critic Learning Rate & $3 \times 10^{-4}$ \\
        Actor Learning Rate & $3 \times 10^{-4}$ \\
        Alpha Learning Rate & $3 \times 10^{-4}$ \\
        Discount Factor ($\gamma$) & 0.99 \\
        Batch Size & 512 \\
        Replay Buffer Size & $1 \times 10^6$ \\
        Gradient Updates per Generation & 40,000 \\
        Reward Scaling Factor & 0.01 \\
        \midrule
        \multicolumn{2}{c}{\textit{Distributional Critic (C51)}} \\
        \midrule
        Architecture & ResNet (2 Blocks, 512 Dim) \\
        Number of Atoms ($N$) & 101 \\
        Support Range ($V_{min}, V_{max}$) & $[-200, 1000]$ \textsuperscript{\textdagger}\\
        Weight Decay & 0.0 \\
        \midrule
        \multicolumn{2}{c}{\textit{Dominated Novelty Search (DNS)}} \\
        \midrule
        Archive Capacity & 250 \\
        DNS Nearest Neighbors ($k$) & 7 \\
        % Selection Metric & Dominated Novelty \\
        Mutation Proportion (GA) & 0.5 \\
        Iso Mutation $\sigma$ & 0.005 \\
        Line Mutation $\sigma$ & 0.1 \\
        Gradient Mutation Steps & 10 \\
        Environment Batch Size & 10 \\
        \bottomrule
    \end{tabular}
    \vspace{0.1cm}
    \footnotesize{
    \newline
    \textsuperscript{\textdagger} \textit{Exceptions: Ant $([-150, 2500])$ and Humanoid $([-150, 1500])$ use expanded ranges to account for higher maximum returns.}}
    \label{table:hyperparams}
\end{table}

\subsection{Impact of Archive Structure (DNS vs. MAP-Elites)} \label{app:me_abl}
To ensure that the performance gains observed in QDHUAC are attributable to the target-free critic architecture rather than the choice of evolutionary archive, we conducted an ablation study substituting Dominated Novelty Search (DNS) with a standard structured MAP-Elites grid with QDHUAC.

\begin{figure*}
    \centering
    \includegraphics[width=0.75\linewidth]{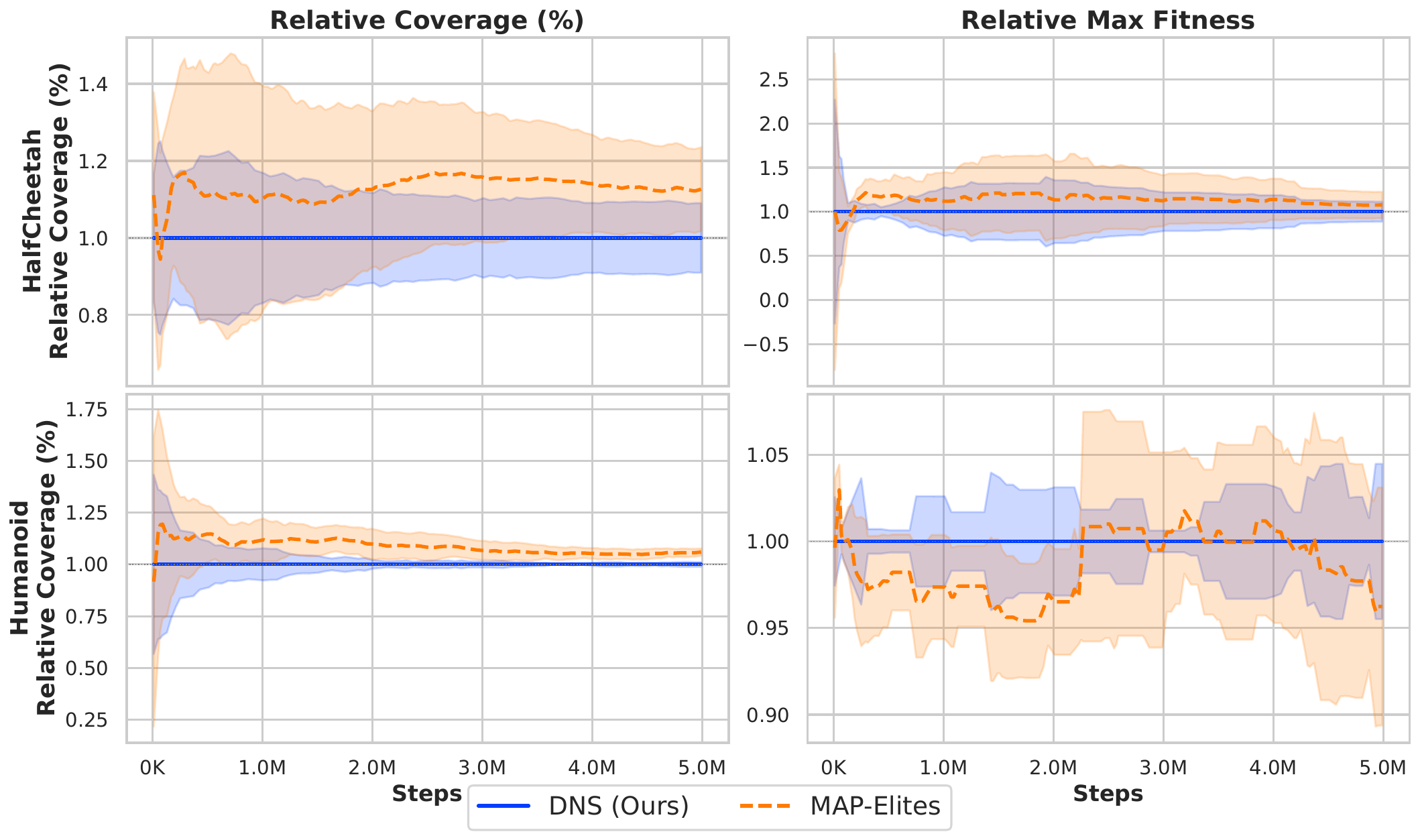}
    \caption{\textit{Evaluating Archive Comparability.} Relative performance of a structured MAP-Elites archive compared against the unstructured DNS archive (normalized to a baseline of 1.0, solid blue line) within the QDHUAC framework. Shaded regions represent the standard deviation across 5 independent seeds. The lack of significant divergence in Max Fitness confirms that the sample efficiency gains of QDHUAC are robust to the choice of underlying evolutionary archive.}
    \label{fig:me_ablation}
\end{figure*}

Figure \ref{fig:me_ablation} illustrates the relative performance of MAP-Elites normalized against DNS (baseline 1.0) on the HalfCheetah and Humanoid environments over 5 million steps. The results demonstrate that the choice of archive does not significantly alter the maximum fitness capacity of the algorithm; the relative Max Fitness heavily overlaps 1.0 with no statistically significant divergence. MAP-Elites exhibits a small relative increase in Coverage, which is expected given its uniformly discretised grid that explicitly forces spatial diversity, whereas the unstructured nature of DNS allows for slightly more aggressive local exploitation.

Crucially, because both archive variants perform comparably and successfully leverage the High-UTD updates, we conclude that the improvements in sample efficiency over baselines (e.g., PGA-ME) are caused by our target-free hybrid normalised critic, rather than the isolated inclusion of DNS. 

\end{document}